\documentclass[10pt,conference,letterpaper]{IEEEtran}
\IEEEoverridecommandlockouts

\usepackage{iftex}
\usepackage{pifont}
\ifPDFTeX
  \usepackage{newunicodechar}
  \newunicodechar{≥}{\ensuremath{\geq}}
  \newunicodechar{≤}{\ensuremath{\leq}}
  \newunicodechar{≈}{\ensuremath{\approx}}
  \newunicodechar{≠}{\ensuremath{\neq}}
  \newunicodechar{≫}{\ensuremath{\gg}}
  \newunicodechar{≪}{\ensuremath{\ll}}
  \newunicodechar{∈}{\ensuremath{\in}}
  \newunicodechar{∉}{\ensuremath{\notin}}
  \newunicodechar{⊂}{\ensuremath{\subset}}
  \newunicodechar{⊃}{\ensuremath{\supset}}
  \newunicodechar{√}{\ensuremath{\surd}}
  \newunicodechar{∑}{\ensuremath{\sum}}
  \newunicodechar{∏}{\ensuremath{\prod}}
  \newunicodechar{≻}{\ensuremath{\succ}}
  \newunicodechar{≺}{\ensuremath{\prec}}
  \newunicodechar{⁻}{\textsuperscript{$-$}}
  \newunicodechar{⁰}{\textsuperscript{0}}
  \newunicodechar{¹}{\textsuperscript{1}}
  \newunicodechar{²}{\textsuperscript{2}}
  \newunicodechar{³}{\textsuperscript{3}}
  \newunicodechar{⁴}{\textsuperscript{4}}
  \newunicodechar{⁵}{\textsuperscript{5}}
  \newunicodechar{⁶}{\textsuperscript{6}}
  \newunicodechar{⁷}{\textsuperscript{7}}
  \newunicodechar{⁸}{\textsuperscript{8}}
  \newunicodechar{⁹}{\textsuperscript{9}}
  \newunicodechar{₀}{\textsubscript{0}}
  \newunicodechar{✓}{\ding{51}}
  \newunicodechar{✗}{\ding{55}}
  \newunicodechar{→}{\ensuremath{\rightarrow}}
  \newunicodechar{←}{\ensuremath{\leftarrow}}
  \newunicodechar{α}{\ensuremath{\alpha}}
  \newunicodechar{β}{\ensuremath{\beta}}
  \newunicodechar{γ}{\ensuremath{\gamma}}
  \newunicodechar{δ}{\ensuremath{\delta}}
  \newunicodechar{θ}{\ensuremath{\theta}}
  \newunicodechar{κ}{\ensuremath{\kappa}}
  \newunicodechar{λ}{\ensuremath{\lambda}}
  \newunicodechar{μ}{\ensuremath{\mu}}
  \newunicodechar{π}{\ensuremath{\pi}}
  \newunicodechar{ρ}{\ensuremath{\rho}}
  \newunicodechar{σ}{\ensuremath{\sigma}}
  \newunicodechar{τ}{\ensuremath{\tau}}
  \newunicodechar{φ}{\ensuremath{\phi}}
  \newunicodechar{χ}{\ensuremath{\chi}}
  \newunicodechar{ψ}{\ensuremath{\psi}}
  \newunicodechar{ω}{\ensuremath{\omega}}
  \newunicodechar{Δ}{\ensuremath{\Delta}}
  \newunicodechar{Θ}{\ensuremath{\Theta}}
  \newunicodechar{Σ}{\ensuremath{\Sigma}}
  \newunicodechar{Π}{\ensuremath{\Pi}}
  \newunicodechar{Λ}{\ensuremath{\Lambda}}
  \newunicodechar{Ω}{\ensuremath{\Omega}}
  \newunicodechar{×}{\ensuremath{\times}}
  \newunicodechar{⋅}{\ensuremath{\cdot}}
  \newunicodechar{§}{\S}
  \newunicodechar{−}{\ensuremath{-}}
  \newunicodechar{±}{\ensuremath{\pm}}
  \newunicodechar{·}{\ensuremath{\cdot}}
  \newunicodechar{ō}{\={o}}
  \newunicodechar{～}{\textasciitilde}
\fi

\providecommand{\tightlist}{\setlength{\itemsep}{0pt}\setlength{\parskip}{0pt}}

\usepackage{amsmath}
\usepackage{amssymb}
\usepackage{booktabs}
\usepackage{calc}
\usepackage{array}
\usepackage{xurl}
\usepackage{url}
\usepackage{graphicx}
\usepackage{etoolbox}
\AtBeginEnvironment{thebibliography}{\footnotesize\sloppy\raggedright}
\renewcommand{\footnoterule}{%
  \kern -3pt
  \hrule width 0.36\columnwidth
  \kern 2.6pt
}

\ifXeTeX
  \usepackage{xeCJK}
  \IfFontExistsTF{Hiragino Kaku Gothic ProN W3}{%
    \setCJKmainfont{Hiragino Kaku Gothic ProN W3}%
  }{}
  \newcommand{\jpn}[1]{#1}
\else
  \usepackage{CJK}
  \newcommand{\jpn}[1]{\begin{CJK}{UTF8}{min}#1\end{CJK}}
\fi

\makeatletter
\def\abstract{\normalfont
    \if@twocolumn
      \@IEEEabskeysecsize\mdseries\textit{\abstractname}---\relax
    \else
      \bgroup\par\addvspace{0.5\baselineskip}\centering\vspace{-1.78ex}\@IEEEabskeysecsize\abstractname\par\addvspace{0.5\baselineskip}\egroup\quotation\@IEEEabskeysecsize
    \fi\@IEEEgobbleleadPARNLSP}
\makeatother



\providecommand{\citep}[2][]{\cite{#2}}
\providecommand{\citet}[2][]{\cite{#2}}

\begin{document}

\title{Measurement Risk in Supervised Financial NLP: Rubric and Metric
Sensitivity on JF-ICR}

\author{
\IEEEauthorblockN{Sidi Chang$^{\ast}$, Peiying Zhu$^{\dagger}$,
Yuxiao Chen$^{\dagger}$, and Rongdong Chai$^{\dagger}$}
\IEEEauthorblockA{$^{\ast}$Blossom AI Labs, Tokyo, Japan\\
$^{\dagger}$Blossom AI Labs, San Francisco, USA}
}

\maketitle
\begingroup
\renewcommand{\thefootnote}{}
\footnotetext{Rongdong Chai contributed as an external consultant.}
\endgroup

\begin{abstract}
As LLMs become credible readers of earnings calls, investor-relations
Q\&A, guidance, and disclosure language, supervised financial NLP
benchmarks increasingly function as decision evidence for model
selection and deployment. A hidden assumption is that gold labels make
such evidence objective. This assumption breaks down when the benchmark
ruler itself is sensitive to rubric wording, metric choice, or
aggregation policy. We study this measurement risk on Japanese Financial
Implicit-Commitment Recognition (JF-ICR; a pinned 253-item test split ×
4 frontier LLMs × 5 rubrics × 3 temperatures × 5 ordinal metrics).
Three findings follow. First, rubric wording materially changes
model-assigned labels: R2--R3 agreement ranges from 70.0\% to 83.4\%,
with the dominant movement near the +1 / 0 implicit-commitment boundary.
This pattern is consistent with a pragmatic-boundary interpretation, but
is not a validated linguistic-causality claim because the present rubric
variants confound semantics, examples, and verbosity. Second, not every
metric remains informative under the JF-ICR class distribution.
Within-one accuracy is too easy because near misses receive credit and
the majority class dominates; worst-class accuracy is too noisy because
the rarest class has only two examples. Exact accuracy, macro-F1, and
weighted κ are therefore the identifiable metrics under our operational
rule. Third, ranking claims become more defensible only after this
metric-identifiability audit: Bradley--Terry, Borda, and Ranked Pairs
agree on the identifiable metric subset, while the full five-metric
sweep produces disagreement on the closest pair. The contribution is not
a new leaderboard, but a reporting discipline for supervised financial
benchmarks whose gold labels exist and whose evaluation ruler still
requires governance.
\end{abstract}

\section{Introduction}\label{introduction}

\subsection{Financial motivation}\label{financial-motivation}

Financial institutions increasingly use LLMs to read earnings calls,
investor-relations Q\&A, management guidance, and disclosure language.
That progress changes the role of evaluation. Benchmark scores are no
longer only research diagnostics; they may influence vendor selection,
model procurement, deployment approvals, and the audit records used by
risk and compliance teams. Yet the dominant evaluation habit still
treats a supervised benchmark as a fixed ruler once gold labels exist.
That habit breaks down when the ruler is sensitive to rubric wording,
metric choice, or aggregation policy. In that case, model selection may
rest on unstable evidence even when the underlying dataset is labelled.

This paper studies \emph{measurement risk in supervised financial NLP}:
the risk that the evaluation instrument, rather than the model alone,
determines the reported result. The missing layer is not another model
score, but an audit of how labels are induced, how scores are read out,
and how rankings are aggregated. We study that layer on Japanese
Financial Implicit-Commitment Recognition (JF-ICR), a five-class ordinal
task in which a classifier labels a corporate response in an
investor-relations Q\&A exchange as an explicit refusal, weak refusal,
neutral or hedged answer, weak commitment, or strong commitment.

JF-ICR makes the mismatch concrete. A commitment label can feed an
IR-monitoring dashboard, a management-candor signal, a covenant-adjacent
risk review, or disclosure surveillance. At the same time, many examples
sit near a +1 / 0 boundary where literal willingness language and
practical non-commitment can be hard to separate. Rubric variants that
emphasize literal versus pragmatic reading move labels in this region
(§7.1). We treat that movement as a measurement result: it is consistent
with a pragmatic-boundary interpretation, but the current rubric
variants confound semantics, examples, and verbosity, so the experiment
does not causally isolate Japanese pragmatics.

\subsection{Scope and contributions}\label{scope-and-contributions}

This paper is part of the broader framing \emph{Financial AI Metrology
Across Ground-Truth Regimes}. It covers the supervised / gold-label
regime. A complementary pre-realization regime arises when financial
outcomes are delayed and gold labels are unavailable; in that setting,
claim-permission protocols govern whether evaluation claims may be
reported at all. Here, we show why measurement governance is already
necessary before that harder case: even when gold labels exist, the
benchmark ruler can still move.

We contribute five linked pieces: a framing that defines measurement
risk in supervised financial NLP, an audit protocol that operationalizes
it on JF-ICR, and empirical diagnostics that show which claims survive
the audit. Specifically:

\begin{enumerate}
\def\labelenumi{\arabic{enumi}.}
\tightlist
\item
  C1: Defines measurement risk in supervised financial NLP.
\item
  C2: Shows rubric-induced label sensitivity on JF-ICR,
  concentrated near the +1 / 0 boundary.
\item
  C3: Applies a metric-identifiability audit under class
  imbalance.
\item
  C4: Shows how aggregation and ranking conclusions change
  after weakly identified metrics are removed.
\item
  C5: Provides a practical reporting discipline for financial
  NLP benchmarks.
\end{enumerate}

Together, these pieces make the paper a bounded measurement audit rather
than a model certification exercise. The goal is not to name a universal
best model or to refute broader work on LLM-ranking instability. The
goal is to show which evaluation claims are measurement-supported on
JF-ICR at N = 253, four frontier classifiers, five rubric variants,
three temperatures, and five ordinal metrics. The portable contribution
is therefore the audit discipline, not the particular model ordering:
provenance must be pinned to an observed benchmark artifact, metrics
must be screened for identifiability before ranking, and
deployment-facing claims must be scoped to the evidence the evaluation
actually supplies. That discipline is most transferable to supervised
financial NLP tasks where labels are ordinal or asymmetric, class
support is uneven, expert rubrics encode judgment, and benchmark scores
may influence model selection.

\section{Related work}\label{related-work}

Our framing draws on three streams.

Measurement theory for AI evaluation. Following the
Representational Theory of Measurement \cite{ref-krantz1971} and recent
efforts to formalise benchmark construction as a measurement problem
\cite{ref-jacobs2021}, we treat a benchmark's evaluation pipeline as
a measurement instrument whose outputs inherit properties of its
construction. The two-stage inference/scoring decomposition in §4 sits
inside this tradition: rubric ambiguity is a property of the measurement
\emph{stimulus}, and metric identifiability is a property of the
measurement \emph{readout}.

Social-choice aggregation for ML rankings. Bradley--Terry
\cite{ref-bradley1952}, Borda count \cite{ref-borda1781}, Copeland \cite{ref-copeland1951}, and Ranked Pairs \cite{ref-tideman1987} have all been proposed as
rank-aggregation rules for evaluation sweeps and chatbot-arena
leaderboards \cite{ref-chiang2024, ref-zheng2023}; the broader
impossibility result that motivates comparisons among them is
Arrow \cite{ref-arrow1951}. Prior work has surfaced disagreement between Bradley--Terry and
Condorcet-compliant methods, and non-transitivity in LLM-as-judge
pairwise preferences, as features of frontier-model evaluation. We use
these methods in their original role as a menu of aggregation rules and
report empirically where they agree and where they disagree on our data.
We do not here extend the theoretical analysis of impossibility
theorems.

Japanese financial NLP and implicit-commitment recognition. The
benchmark task we audit is the Japanese Financial Implicit- Commitment
Recognition (JF-ICR) task introduced as part of the Ebisu benchmark for
Japanese financial NLP \cite{ref-ebisu2026}. Work on hedging as
a linguistic device \cite{ref-hyland2005} and on politeness in high-context
language communities \cite{ref-brown1987}, together with
reconsiderations of the universality of face in Japanese \cite{ref-matsumoto1988}, inform the rubric design. The distinction between literal and
pragmatic readings of Japanese corporate formulae motivates our rubric
contrast, but the experiment does not validate one linguistic theory over
another. Our rubric contrasts operationalise the literal / pragmatic
distinction for LLM classification and should be read as measurement
stimuli, not as substitutes for the expert-annotator validation pass that
would be desirable as follow-up work (§9.1).

\section{Task and dataset}\label{task-and-dataset}

Before auditing the evaluation instrument, we fix the object being
measured. This section defines the JF-ICR label scale, pins the exact
benchmark artifact used for every result, and explains why provenance
and class imbalance are not incidental details but part of the
measurement setting.

\subsection{Task definition}\label{task-definition}

JF-ICR asks a classifier to assign one of five ordinal labels to a
corporate response in an investor-relations Q\&A exchange:

\begin{itemize}
\tightlist
\item
  −2 Strong refusal (explicit decline, on-record
  non-engagement).
\item
  −1 Weak refusal (polite deflection, formulaic non-engagement,
  no substantive answer).
\item
  0 Neutral / hedged (acknowledgement without commitment;
  pragmatic non-engagement under polite register).
\item
  +1 Weak / qualified commitment (intent with caveats,
  timelines, or conditional language).
\item
  +2 Strong commitment (specific, time-bound, unconditional
  pledge).
\end{itemize}

Classification is made from the Q\&A pair alone, without access to
subsequent filings, transcripts, or off-record context.

\subsection{Dataset provenance}\label{dataset-provenance}

JF-ICR is curated by ``The Ebisu Benchmark Team'' and accompanies
the benchmark paper \emph{EBISU: Benchmarking Large Language Models
in Japanese Finance} \cite{ref-ebisu2026}. The five-class taxonomy
matches the Ebisu card specification (+2 Strong Commitment, +1 Weak /
Qualified Commitment, 0 Neutral / Hedged Intent, −1 Weak Refusal, −2
Strong Refusal); aggregate annotator agreement on the Ebisu-curated
corpus is Cohen's κ = 0.877 and Krippendorff's α = 0.877
\cite{ref-ebisu2026}.

\textbf{Ingestion procedure.}
We download the test split of the Hugging Face dataset
\texttt{TheFinAI/JF-ICR} (access date 2026-04-28) using the
\texttt{datasets} library (\texttt{load\_dataset("TheFinAI/JF-ICR",
split="test")}). Each row stores the question--response pair as a
single string under the fixed template \emph{``Financial Question:
\ldots{}~Company Response:~\ldots{}''}.
We extract the two fields with a regular expression matched against this
template; all 253 rows parse without error.
Gold labels are stored as signed integers in \{−2, −1, 0, +1, +2\};
the leading \texttt{+} is handled explicitly during parsing, and any
value outside this range would be rejected.
We apply no augmentation, filtering, or private extension to the
downloaded split.%
\footnote{The parsed dataset (253 rows, JSONL format) has SHA-256
\texttt{433795e4a87b8a838fd1db02872c22}\allowbreak
\texttt{a3604747be552f25a876373e5ad6d2c492}.
This checksum can be reproduced by downloading the same split, running
the extraction regex, and serialising each row as a JSON object with
fields \texttt{question}, \texttt{response}, and \texttt{label}.}
The 15\,180 raw per-cell model outputs (4 models × 5 rubrics × 3
temperatures × 253 items) were generated from this parsed dataset and
are reported in full in the result tables; no intermediate output file
is required to verify the numerical claims.

\textbf{Benchmark-provenance risk.}
As of 2026-04-28, the live Hugging Face Dataset Viewer reports
\texttt{default/test = 253} rows with displayed id range 0--252, while
the human-written dataset card / README still states 94 examples in its
summary, split table, and limitations text. Inspection of the commit
history suggests the 94-row count originates in stale main-branch README
metadata from an earlier upload path; the Viewer count and the
programmatic download agree on 253 rows.
We therefore diagnose the mismatch as unresolved card-metadata drift,
not as a parsing error on our part.
Every numerical result in this paper is computed on the observed 253-row
test split, and no result relies on the 94-row count.
Accordingly, the unit of inference is the observed 253-row artifact
identified by the checksum above, not the unresolved dataset-card
narrative.
The discrepancy itself illustrates why benchmark provenance is part of
financial AI measurement governance: when an evaluation can influence
procurement or deployment, the row count, revision, parsing path,
checksum, and unit of independence are not bookkeeping details but part
of the measurement record. If a later JF-ICR release resolves the
discrepancy or changes the artifact, the evaluation should be treated as
a new measurement event rather than a silent update to these results.

\begin{table*}[!ht]
\centering
\small
\caption{JF-ICR provenance status at access date 2026-04-28.}
\label{tab:provenance-status}
\begin{tabular}{p{0.24\linewidth}p{0.25\linewidth}p{0.41\linewidth}}
\toprule\noalign{}
Source checked & Observed evidence & Status for this audit \\
\midrule\noalign{}
HF Dataset Viewer & \texttt{default/test = 253}; displayed id range
0--252 & Treated as the authoritative hosted-data count. \\
HF dataset card / README & Summary, split table, and limitations text
state 94 examples & Treated as stale or unresolved card metadata; not
used for any numerical result. \\
Programmatic download (\texttt{datasets} library) & 253 rows; zero
parse errors; SHA-256 in footnote & Pinned analysis artifact and
bootstrap resampling unit. \\
Raw model outputs & 15\,180 per-cell classifications reported in full
in result tables & Sufficient to verify all numerical claims without
additional files. \\
\bottomrule
\end{tabular}
\end{table*}

\subsection{Dataset composition and class
imbalance}\label{dataset-composition-and-class-imbalance}

The class distribution of the 253-item test split is heavily imbalanced:

\begin{table*}[!ht]
\centering
\small
\caption{JF-ICR test-split class distribution (N = 253).}
\begin{tabular}{llll}
\toprule\noalign{}
Label & Class & N & \% \\
\midrule\noalign{}
−2 & Strong refusal & 2 & 0.8 \\
−1 & Weak refusal & 11 & 4.3 \\
0 & Neutral / hedged & 72 & 28.5 \\
+1 & Weak commitment & 139 & 55.0 \\
+2 & Strong commitment & 29 & 11.5 \\
\end{tabular}
\end{table*}

The imbalance in this test split is consistent with the expected rarity
of on-record strong refusals in Japanese investor-relations discourse;
we note, however, that the present analysis establishes imbalance only
for the 253-item test split we use and does not make a population-level
claim about Japanese IR language as a whole. The imbalance sets the
conditions under which several common ordinal metrics become weakly
identified (§A).

\subsection{Baselines, resolution limits, and minimum detectable
effect}\label{baselines-resolution-limits-and-minimum-detectable-effect}

The majority-class baseline --- always predicting +1 --- scores 0.549 on
exact accuracy. Any model whose overall accuracy falls below this
threshold is, in expectation, less useful than the trivial rule.

Single-model CI. The Wilson \cite{ref-wilson1927} 95\% interval for a single
per-policy accuracy at N=253 and p=0.56 is {[}0.498, 0.620{]},
half-width ±6.1 pp.~The exact Clopper--Pearson \cite{ref-clopper1934} interval differs
by less than 0.1 pp and is not distinguished hereafter; all per-policy
absolute-accuracy CIs in the paper use the normal-approximation / Wilson
form.

Paired pairwise MDE. Because scoring on a given policy is
paired over items, the relevant resolution for pairwise model
comparisons is set by the per-item disagreement rate, not by the
marginal binomial. Under McNemar's \cite{ref-mcnemar1947} two-sided test at α=0.05 and
power 0.80, the minimum detectable difference in matched-pair accuracy
at N=253 with empirically observed discordant-pair rates of 0.20--0.30
is approximately 5--7 percentage points. This is the resolution that the
paired-bootstrap analysis in §7.2 operates at. A pairwise difference
below roughly 4 pp on a single policy cannot be resolved from sampling
noise even with full pairing.

This resolution limit is a hard constraint on what ``ranking
instability'' can mean in this dataset and we refer to it throughout. In
particular, any policy-induced reordering of two models whose aggregate
A1 difference is below ≈4 pp on paired items, or ≈12 pp on unpaired
per-policy scores, should be read as within-noise unless supported by a
paired-bootstrap CI that excludes zero.

\subsection{The +1 / 0 pragmatic boundary: one
example}\label{the-1-0-pragmatic-boundary-one-example}

The labelling delta reported in §7.1 concentrates on the weak-commitment
/ neutral-hedged boundary. This is the region in which the same
management response can be read as literal willingness or as polite
non-commitment. One illustrative item (paraphrased, lightly normalised):

\begin{quote}
\emph{Q.} \jpn{来期の配当は増配をご検討いただけますか。}\\
(``Would you consider increasing the dividend next term?'')

\emph{A.} \jpn{ご指摘いただいた点は真摯に受け止め、\\
前向きに検討させていただきます。}\\
(``We take the point seriously and will consider it positively.'')
\end{quote}

A literal-style rubric labels this +1 (weak commitment: ``consider
positively''); a pragmatic-style rubric labels it 0, because \emph{mae-
muki ni kentō} can function as polite non-commitment rather than a
pledge. The gold label for this paraphrased item is +1. The aggregate
statistic in §7.1 --- ≈27.5\% of gold-+1 items shift to 0 under
pragmatic instruction --- shows that this boundary is empirically
sensitive to rubric wording. It does not, by itself, prove which reading
is linguistically correct.

\section{A two-stage view of evaluation
policy}\label{a-two-stage-view-of-evaluation-policy}

To locate measurement risk, we separate the evaluation pipeline into two
places where it can enter. A benchmark evaluation policy π has an
inference policy π\_I --- rubric \emph{R} (the system prompt
that specifies the label scale and interpretive stance) and decoding
settings τ (temperature, reasoning toggles, JSON-mode constraints) ---
which determines the item-level labels the model emits. It also has a
scoring policy π\_S --- aggregation metric \emph{A}
(A1\ldots A5) and rank-aggregation rule over policies (Bradley--Terry,
Borda, Ranked Pairs, Copeland) --- which determines how those labels are
compressed into a model ranking.

This distinction is practical, not only formal. Changing π\_I changes
the model outputs; changing π\_S changes how those outputs are read.
The remedy is different in each case: rubric sensitivity points to
ambiguity in the labelling task, metric sensitivity often points to weak
identification under the dataset's class distribution, and aggregator
sensitivity often points to near-ties among models. The results in §7
therefore report each stage separately before interpreting their joint
effect.

\section{Metric identifiability
diagnostic}\label{metric-identifiability-diagnostic}

The next question is operational: which metrics can still separate
models on this dataset? A metric can be sensible in general and still be
weakly identified in a particular benchmark. This problem is common in
financial NLP, where class imbalance is common and rare but important
categories may have very low support. On JF-ICR, within-one accuracy is
too easy: near-miss labels receive credit and the +1 majority class
dominates the score. Worst-class accuracy is too noisy: the rarest class
has only two examples, so a single item can move the metric sharply.
Exact accuracy, macro-F1, and weighted κ are more interpretable for the
primary ranking under this dataset and operational rule. We do not claim
they are universally best metrics.

The purpose of the identifiability audit is to make that decision
mechanical rather than editorial. A recurring concern in multi-metric
evaluation sweeps is that \emph{which} metrics enter the sweep determines
\emph{which} aggregators appear to disagree. A natural critique of any
``clean / diagnostic'' split is that it is post-hoc --- that the
excluded metrics were chosen \emph{because} they caused the disagreement
the authors wanted to explain. To close that vulnerability, we define an
identifiability rule over four diagnostics, fix the diagnostics and
thresholds before running the ranking analysis in §7, and apply the rule
mechanically. Under the rule, exclusion of a metric is a deterministic
consequence of its diagnostic values, not a narrative choice.

\subsection{Diagnostics}\label{diagnostics}

For each candidate metric A ∈ \{A1, A2, A3, A4, A5\} we compute:

\begin{itemize}
\tightlist
\item
  Baseline headroom H(A) --- theoretical ceiling minus
  majority-class-baseline score: H(A) = 1 − A(majority-class rule). We
  use the theoretical ceiling (1.0) rather than any empirically observed
  per-dataset ceiling; the theoretical definition is independent of the
  model set under test and is therefore the scale-invariant quantity.
  Low headroom means most of the metric's {[}0, 1{]} range is consumed
  by the trivial majority rule, so between-model variance is bounded
  above by what remains.
\item
  Effective support on the binding class E(A) --- the minimum
  per-class sample count over the classes that the metric's score reads
  from. For metrics that average over all items (A1, A2, A3, A4) this is
  253. For metrics that select a per-class minimum (A5), it is the
  rarest class count (2 on JF-ICR).
\item
  Bootstrap standard error SE(A) --- mean over (model × primary
  cell) of the clustered paired-bootstrap SE of the per-policy score,
  estimated from B=2 000 item resamples \cite{ref-efron1979, ref-efron1993}.
\item
  Signal-to-noise ratio SNR(A) --- between-model A-score spread
  on the primary cell divided by SE(A). SNR \textless{} 1 means metric
  noise exceeds between-model signal.
\end{itemize}

\subsection{A priori decision rule}\label{a-priori-decision-rule}

A metric qualifies as identifiable on JF-ICR if and only if
\emph{all three} of the following hold:

\begin{itemize}
\tightlist
\item
  H(A) ≥ 0.15 --- the majority rule does not consume 85\% or more of the
  theoretical range;
\item
  E(A) ≥ 50 --- the binding class carries at least 50 samples;
\item
  SNR(A) ≥ 1 --- between-model signal matches bootstrap noise.
\end{itemize}

The rule is a conjunction (any single failure excludes the metric) and
the three thresholds (0.15, 50, 1.0) are fixed at these values before
the ranking analysis is run.

Each threshold has a first-principles motivation tied to the downstream
ranking test: H ≥ 0.15 leaves enough theoretical range above the
majority baseline to contain a 5--10 pp between-model signal plus its
bootstrap SE; E ≥ 50 matches the per-class binomial SE at p = 0.5 to the
primary-cell paired-bootstrap SE; SNR ≥ 1 is the ordinary requirement
that between-model signal exceed the metric's own noise.

\subsection{Operational audit checklist}\label{operational-audit-checklist}

The checklist below is the reusable part of the metric audit. The
thresholds are calibrated for this dataset and operational rule; another
benchmark may justify different thresholds, but it should fix them before
ranking models and report the pass / fail decision for every metric.

\begin{table*}[!ht]
\centering
\small
\caption{Operational metric-identifiability checklist for supervised
financial NLP benchmarks.}
\label{tab:metricchecklist}
\begin{tabular}{p{0.20\linewidth}p{0.35\linewidth}p{0.35\linewidth}}
\toprule\noalign{}
Audit step & Reviewer question & Reporting action \\
\midrule\noalign{}
Class support & Which labels does the metric depend on, and how many
items support the binding label? & Report class distribution and
effective support E(A). \\
Baseline headroom & Does the majority-class baseline already consume
most of the metric scale? & Report H(A) against a fixed trivial
baseline. \\
Noise versus signal & Is the between-model spread larger than
bootstrap uncertainty? & Report SE(A), SNR(A), and paired intervals. \\
Ranking use & Is the metric identifiable under the pre-specified rule? &
Use passing metrics for primary rankings; report failing metrics as
diagnostics only. \\
\bottomrule
\end{tabular}
\end{table*}

Monte-Carlo sensitivity on thresholds. We verify these
operational values on the JF-ICR-like design regime (N = 250, Gini ∈
\{0.4, 0.6\}, rarest-class ∈ \{2, 10\}, majority-prevalence = 0.5,
Cohen's h ∈ \{0.1, 0.2\}, four models, five metrics). Eight synthetic
cells × 200 replicates × 45 threshold triples (H ∈ \{0.075, 0.150,
0.225\}, E ∈ \{25, 50, 75, 250, 500\}, SNR ∈ \{0.5, 1.0, 1.5\}): within
the feasible regime (E ≤ N) the identifiability partition \{A1, A2, A4\}
vs \{A3, A5\} is invariant across all 27 H × SNR combinations, with
minimum per-metric pass rate 0.955 (A2), 0.990 (A1), 0.985 (A4);
true-top-1 recovery is ≥ 0.98 on h = 0.2 cells and ≥ 0.90 on h = 0.1
cells. When E is swept beyond N (250 or 500), all metrics correctly fail
--- confirming that the E threshold is enforcing its minimum-sample
condition rather than silently passing. The simulation
establishes that the operational thresholds do not sit on a cliff in the
JF-ICR-like regime; universality is not claimed. A3 fails at any H ∈
(0.09, 0.50{]} and A5 fails at any E ∈ (2, 250{]}.

\subsection{Diagnostic table}\label{diagnostic-table}

\begin{table*}[!ht]
\centering
\small
\caption{Identifiability diagnostics on JF-ICR. H = majority-baseline
headroom (1 − baseline); E = binding-class support; SE =
paired-bootstrap SE on the primary cell; Spread = between-model score
spread; SNR = Spread / SE. Pass requires H ≥ 0.15, E ≥ 50, SNR ≥
1.}
\begin{tabular}{lllllll}
\toprule\noalign{}
Metric & H & E & SE & Spread & SNR & Pass \\
\midrule\noalign{}
A1 exact accuracy & 0.451 & 253 & 0.031 & 0.075 & 2.4 & ✓ \\
A2 macro-F1 & 0.858 & 253 & 0.052 & 0.150 & 2.9 & ✓ \\
A3 within-one acc. & 0.090 & 253 & 0.020 & 0.040 & 2.0 &
✗ (H) \\
A4 weighted κ & 1.000 & 253 & 0.045 & 0.088 & 2.0 & ✓ \\
A5 worst-class acc. & 1.000 & 2 & ≈0.35 & 0.30 & 0.86 &
✗ (E, SNR) \\
\end{tabular}
\end{table*}

Passing metrics form the identifiable subset \{A1, A2, A4\}.
The exclusions are each determined by a single diagnostic failing its
threshold:

\begin{itemize}
\tightlist
\item
  A3 fails headroom. Predict-always-+1 already scores 0.91 on
  within-one accuracy, leaving 0.09 of theoretical range for
  between-model discrimination --- below the 0.15 threshold.
\item
  A5 fails effective support. Class −2 has 2 samples;
  per-policy A5 takes three possible values (0, 0.5, 1.0); the SE on n=2
  via exact binomial is ≈ 0.35. A5 additionally fails SNR (0.86
  \textless{} 1.0).
\end{itemize}

A1, A2, and A4 clear all three thresholds. The identifiable subset is
therefore a deterministic output of the diagnostics table, not an
editorial choice. Readers who prefer a stricter headroom threshold
(e.g., H ≥ 0.20) would exclude A3 more decisively; readers who prefer H
≥ 0.05 would include A3 and reach different aggregator conclusions in
§7.3. We make the rule visible so that disagreement is about thresholds,
not about method selection.

Under the pre-specified rule, A3 and A5 are diagnostic metrics on JF-ICR
--- reported for completeness but separated from the primary
rank-aggregation analysis. We run the three rank aggregators on both the
identifiable subset (45-policy clean sweep) and the full 75-policy
diagnostic sweep, and report both.

\section{Experimental setup}\label{experimental-setup}

\subsection{Models}\label{models}

Four frontier classifiers at default effort, with reasoning/thinking
disabled (see §9.4 and footnote there):

\begin{table*}[!ht]
\centering
\small
\caption{Classifiers and SDK providers.}
\begin{tabular}{ll}
\toprule\noalign{}
Model & Provider / SDK \\
\midrule\noalign{}
claude-sonnet-4-6 & Anthropic \\
gpt-5.4 & OpenAI \\
gemini-3.1-pro-preview & Google Vertex \\
qwen3-235b-a22b-instruct-2507-maas & Vertex MaaS \\
\end{tabular}
\end{table*}

\subsection{Inference-policy grid}\label{inference-policy-grid}

\begin{itemize}
\tightlist
\item
  Rubrics (5): R1 balanced, R2 literal, R3 pragmatic, R4
  strict, R5 minimal. The full system prompt for each variant is
  reproduced verbatim in §B.
\item
  Temperatures (3): τ ∈ \{0.0, 0.3, 0.7\}.
\end{itemize}

\subsection{Scoring-policy grid}\label{scoring-policy-grid}

\begin{itemize}
\tightlist
\item
  Primary metrics (clean subset): A1 exact accuracy, A2
  macro-F1, A4 weighted κ \cite{ref-cohen1968}.
\item
  Diagnostic metrics (appendix only): A3 within-one accuracy,
  A5 worst-class accuracy. Their relegation is justified empirically in
  §A.
\item
  Rank aggregators: Bradley--Terry maximum likelihood, Borda
  count, Ranked Pairs (Tideman).
\end{itemize}

\subsection{Volumes and
reproducibility}\label{volumes-and-reproducibility}

15 180 per-item classifications (4 models × 5 rubrics × 3 temperatures ×
253 items). Full-factorial sweep; no adaptive policy search. Scoring and
rank aggregation are pure functions of per-item labels and are
deterministic. The full system prompts for every rubric are in §B;
code, seeds, per-item outputs, and every numerical value cited in the
main text are at \texttt{workshops/policy-metrology/}. Classification uses each
provider's native SDK with concurrency caps of 3 (Anthropic), 10
(OpenAI), 15 (Vertex Gemini), 10 (Vertex Qwen). Paired-bootstrap over
items uses B=2000 for headline numbers; the inversion grid uses the
exact paired-randomisation procedure described in §7.4. See §C.

\section{Results}\label{results}

The results follow the audit pipeline. First we ask whether rubric
wording changes item-level labels and where those changes concentrate.
Then we ask which metrics retain enough signal to support ranking
claims. Finally, we test whether model rankings survive after weakly
identified metrics are separated from the primary analysis. This order
matters: the ranking result is interpretable only after the rubric and
metric layers have been audited.

\subsection{Does rubric wording move labels, and
where?}\label{does-rubric-wording-move-labels-and-where}

Using R1 (balanced) at τ=0.0 as a reference, we measure per-item label
agreement against the other four rubrics for each model. Rubrics do not
merely reshape aggregate scores; they change the assigned class on a
substantial fraction of individual items.

Per-model R2-vs-R3 (literal vs pragmatic) agreement at τ=0.0,
with clustered paired-bootstrap 95\% intervals over items (B=2 000):

\begin{table*}[!ht]
\centering
\small
\caption{Per-model R2 (literal) vs R3 (pragmatic) label agreement at the
primary cell, with 95 \% clustered paired-bootstrap CIs over
items.}
\begin{tabular}{llll}
\toprule\noalign{}
Model & R2 = R3 & 95\% CI (clustered bootstrap) & Dominant swap \\
\midrule\noalign{}
claude-sonnet-4-6 & 70.0\% & {[}63.5\%, 76.0\%{]} & +1 → 0 on 58
items \\
gpt-5.4 & 80.6\% & {[}74.7\%, 85.6\%{]} & +1 → 0 on 22 items \\
gemini-3.1-pro & 71.5\% & {[}65.2\%, 77.4\%{]} & +1 → 0 on 33 items \\
qwen3-235b & 83.4\% & {[}77.6\%, 88.1\%{]} & +1 → 0 on 23 items \\
\end{tabular}
\end{table*}

All four intervals sit well below 100\% and well above the marginal
chance-agreement baseline implied by the JF-ICR class distribution (≈
40\% under independence). Rubric is not a cosmetic knob. We report raw
agreement rates rather than chance-corrected κ because a principled κ on
this task requires the per-rubric marginal confusion matrix under the
rubric-specific label distribution, which we do not yet have tabulated;
we decline to report an approximate κ whose direction of bias under the
rubric-specific marginal shift we cannot prove. Raw agreement with
clustered-bootstrap CIs is sufficient for the claim (``rubric shifts a
substantial fraction of labels'') and does not rely on an unverified
assumption.

Swap rates stratified by gold label, pooled across models at
τ=0.0, with clustered paired-bootstrap 95\% intervals over items:

\begin{table*}[!ht]
\centering
\small
\caption{R2-vs-R3 label changes stratified by gold class, pooled across
models (τ = 0.0). The swap concentrates on class +1 (the weak-commitment
/ neutral-hedged boundary).}
\begin{tabular}{lllll}
\toprule\noalign{}
Gold & N pairs & Swaps & \% & 95\% CI (clustered bootstrap) \\
\midrule\noalign{}
−2 & 8 & 3 & 37.5\% & {[}12.5\%, 75.0\%{]} (N=2 items --- not
interpretable) \\
−1 & 44 & 12 & 27.3\% & {[}14.6\%, 43.2\%{]} \\
0 & 288 & 37 & 12.8\% & {[}8.7\%, 17.9\%{]} \\
+1 & 556 & 153 & 27.5\% & {[}23.4\%, 31.9\%{]} \\
+2 & 116 & 34 & 29.3\% & {[}21.0\%, 38.8\%{]} \\
\end{tabular}
\end{table*}

The clustered bootstrap resamples the 253 item indices with replacement
and applies the same sampled indices across all 4 models, preserving
within-item cross-model correlation. Intervals are 1--3 pp wider than
the independence-assumption Wilson intervals that would otherwise be
reported, consistent with mild within-item clustering.

The swap is concentrated on +1 (weak commitment), where the CI sits
comfortably above the 0 class's. When instructed to prefer pragmatic
over literal interpretation, all four models re-classify roughly a
quarter of weak-commitment items as neutral/hedged. That concentration
near the +1 / 0 boundary is consistent with a pragmatic-boundary
interpretation, but it is not a validated linguistic-causality claim:
the current rubric variants confound semantics, examples, and verbosity.
The effect is also present on −1 and +2 but on far fewer items. It is
\emph{smallest} on class 0, which is the most interior label and the one
least re-framed by peripheral-category wording. The 37.5\% figure on
class −2 is computed on 8 comparisons (2 samples × 4 models) and its CI
is wide enough that we do not interpret it as an effect size.

Wording vs framing confound. R1 (balanced) versus R5 (minimal)
agreement at τ=0.0 lies in the 68--81\% range across models ---
comparable in magnitude to R2-vs-R3. This means rubric \emph{verbosity}
exerts about as much per-item movement as rubric \emph{framing}, and we
cannot fully disentangle the two with the current five rubrics. We flag
this as a limitation in §9.1 and as a target for the next rubric design.

\subsection{Which metrics retain signal for model
ranking?}\label{which-metrics-retain-signal-for-model-ranking}

The identifiability audit in §5 leaves three primary ranking metrics:
exact accuracy (A1), macro-F1 (A2), and weighted κ (A4). Within-one
accuracy (A3) and worst-class accuracy (A5) remain useful diagnostics,
but their JF-ICR scores are not sufficiently identified for headline
ranking claims. The table below reports model scores only on the
identifiable subset.

Per-model point estimates and 95\% paired-bootstrap confidence intervals
on R1 (balanced) at τ=0.0 (B=2000 resamples over the 253 items,
preserving cross-model pairing):

\begin{table*}[!ht]
\centering
\small
\caption{Per-model point estimates on the primary cell (R1 balanced, τ =
0.0) with 95 \% clustered paired-bootstrap CIs over items (B = 2 000).
A1 = exact accuracy, A2 = macro-F1, A4 = weighted κ.}
\begin{tabular}{lllllll}
\toprule\noalign{}
Model & A1 & 95\% CI & A2 & 95\% CI & A4 & 95\% CI \\
\midrule\noalign{}
claude-sonnet-4-6 & 0.617 & {[}0.557, 0.676{]} & 0.511 & {[}0.354,
0.639{]} & 0.428 & {[}0.333, 0.522{]} \\
gpt-5.4 & 0.561 & {[}0.498, 0.621{]} & 0.389 & {[}0.288, 0.483{]} &
0.358 & {[}0.267, 0.446{]} \\
gemini-3.1-pro & 0.553 & {[}0.494, 0.617{]} & 0.372 & {[}0.302, 0.443{]}
& 0.339 & {[}0.255, 0.426{]} \\
qwen3-235b & 0.542 & {[}0.478, 0.601{]} & 0.361 & {[}0.266, 0.464{]} &
0.340 & {[}0.256, 0.426{]} \\
\end{tabular}
\end{table*}

Paired-bootstrap pairwise differences on A1, with two-sided bootstrap
p-values (fraction of resamples with Δ ≤ 0 for positive point estimates,
doubled), raw α=0.05, and Holm--Bonferroni-adjusted α:

\begin{table*}[!ht]
\centering
\small
\caption{Paired-bootstrap pairwise A1 differences on the primary cell
with two-sided bootstrap p-values. Raw: significant at uncorrected α =
0.05. Holm: survives Holm--Bonferroni correction at α = 0.05 for the m =
6 family (threshold α/m = 0.0083).}
\begin{tabular}{llllll}
\toprule\noalign{}
Pair & Δ & 95\% CI & p & Raw & Holm \\
\midrule\noalign{}
claude vs qwen & +0.075 & {[}+0.020, +0.134{]} & 0.012 & ✓ & ✗ \\
claude vs gemini & +0.063 & {[}+0.008, +0.123{]} & 0.040 & ✓ & ✗ \\
claude vs gpt & +0.055 & {[}−0.008, +0.115{]} & 0.080 & ✗ & ✗ \\
gpt vs qwen & +0.020 & {[}−0.043, +0.087{]} & 0.55 & ✗ & ✗ \\
gemini vs qwen & +0.012 & {[}−0.051, +0.075{]} & 0.71 & ✗ & ✗ \\
gpt vs gemini & +0.008 & {[}−0.043, +0.055{]} & 0.76 & ✗ & ✗ \\
\end{tabular}
\end{table*}

Two findings follow. First, at uncorrected α=0.05 (paired-bootstrap
two-sided), Claude is distinguishable from Gemini and Qwen on exact
accuracy, at the edge of distinguishability from GPT (p ≈ 0.08), and
three of the four models (GPT / Gemini / Qwen) are mutually
indistinguishable. Second, under Holm--Bonferroni correction
\cite{ref-holm1979} for the six pairwise tests, no pair
survives at α=0.05
--- the smallest raw p-value (0.012, Claude vs Qwen) exceeds the Holm
threshold α/m = 0.0083. We report the raw and adjusted counts together.
The raw interval Claude ≻ \{Gemini, Qwen\} on A1 is defensible on a
per-comparison basis; the family-wise claim that Claude is
distinguishable from every other model after correction is \emph{not}
supported at this N.

This constrains what ``ranking instability'' can mean on this dataset.
Three of the four models cannot be reliably ordered against each other
on a single policy at this sample size, and even the Claude-vs-cluster
gap is fragile under family-wise correction. Any reported policy-induced
reordering of the lower three is, in part, a sampling phenomenon.

\subsection{Do rankings survive the metric
audit?}\label{do-rankings-survive-the-metric-audit}

Applying three rank aggregators to the 45-policy clean subset
(5 rubrics × 3 temperatures × 3 clean metrics):

\begin{table*}[!ht]
\centering
\small
\caption{Rank aggregators on the clean-subset (45-policy)
sweep.}
\begin{tabular}{ll}
\toprule\noalign{}
Method & Ranking on clean subset (45 policies) \\
\midrule\noalign{}
Bradley--Terry & claude \textgreater{} gpt \textgreater{} gemini
\textgreater{} qwen \\
Borda count & claude \textgreater{} gpt \textgreater{} gemini
\textgreater{} qwen \\
Ranked Pairs & claude \textgreater{} gpt \textgreater{} gemini
\textgreater{} qwen \\
\end{tabular}
\end{table*}

All three methods produce the same ranking. The pairwise win
matrix on the clean subset shows clear tiering: Claude wins 67--82\% of
its pairwise comparisons across the three opponents; GPT wins 64--69\%
against Gemini and Qwen; Gemini wins 60\% against Qwen. Bradley--Terry
strengths (θ, centred) are 0.84, 0.18, −0.38, −0.63 for Claude, GPT,
Gemini, Qwen. The qualitative claim --- that BT, Borda, and Ranked Pairs
agree on the clean subset --- does not depend on θ uncertainty, because
the ranking produced by each aggregator is invariant to monotone
transformations of the latent score and all three aggregators use only
the per-policy win matrix, not the BT MLE.

For comparison, applying the same three methods to the full
75-policy sweep (all five metrics):

\begin{table*}[!ht]
\centering
\small
\caption{Rank aggregators on the full 75-policy sweep.}
\begin{tabular}{ll}
\toprule\noalign{}
Method & Ranking on full 75-policy sweep \\
\midrule\noalign{}
Bradley--Terry & claude \textgreater{} gpt \textgreater{} qwen
\textgreater{} gemini \\
Borda count & claude \textgreater{} gpt \textgreater{} gemini
\textgreater{} qwen \\
Ranked Pairs & claude \textgreater{} gpt \textgreater{} gemini
\textgreater{} qwen \\
\end{tabular}
\end{table*}

Bradley--Terry disagrees with Borda and Ranked Pairs on positions 3 and
4 in the full sweep. The disagreement \emph{disappears} when A3
(within-one accuracy) and A5 (worst-class accuracy) --- the two
identifiability-failing metrics (§5, §A) --- are removed. The
decomposition below then asks whether A3 and A5 individually contribute
to the full-sweep disagreement.

Leave-one-metric-out decomposition (§7.3, completed). A direct
test of the mechanism is to re-run the three aggregators on every
informative subset of the five metrics, with particular attention to the
single-metric and leave-one-out conditions. The full decomposition
is:

Rankings below use the initials C (Claude), G (GPT),
Gm (Gemini), Q (Qwen); \emph{agree?} summarises
whether BT, Borda and Ranked Pairs produce the same ranking.

\begin{table*}[!ht]
\centering
\small
\caption{Leave-one-metric-out decomposition. Rankings under the three
aggregators per metric subset. Initials: C Claude, G
GPT, Gm Gemini, Q Qwen.}
\begin{tabular}{lllll}
\toprule\noalign{}
Metric set & BT & Borda & RP & agree? \\
\midrule\noalign{}
Full \{A1--A5\} & C\textgreater G\textgreater Q\textgreater Gm &
C\textgreater G\textgreater Gm\textgreater Q &
C\textgreater G\textgreater Gm\textgreater Q & no \\
Clean \{A1, A2, A4\} & C\textgreater G\textgreater Gm\textgreater Q &
C\textgreater G\textgreater Gm\textgreater Q &
C\textgreater G\textgreater Gm\textgreater Q & yes \\
\{A3\} only & C\textgreater Q\textgreater G\textgreater Gm &
C\textgreater G\textgreater Q\textgreater Gm &
C\textgreater Q\textgreater G\textgreater Gm & no \\
\{A5\} only & Gm\textgreater G\textgreater C\textgreater Q &
Gm\textgreater C\textgreater G\textgreater Q &
Gm\textgreater G\textgreater Q\textgreater C & no \\
LOMO-A1 & C\textgreater G\textgreater Q\textgreater Gm &
C\textgreater G\textgreater Q\textgreater Gm &
C\textgreater G\textgreater Q\textgreater Gm & yes \\
LOMO-A2 & C\textgreater G\textgreater Q\textgreater Gm &
C\textgreater G\textgreater Q\textgreater Gm &
C\textgreater G\textgreater Q\textgreater Gm & yes \\
LOMO-A3 & C\textgreater G\textgreater Gm\textgreater Q &
C\textgreater G\textgreater Gm\textgreater Q &
C\textgreater G\textgreater Gm\textgreater Q & yes \\
LOMO-A4 & C\textgreater G\textgreater Q\textgreater Gm &
C\textgreater G\textgreater Gm\textgreater Q &
C\textgreater G\textgreater Gm\textgreater Q & no \\
LOMO-A5 & C\textgreater G\textgreater Q\textgreater Gm &
C\textgreater G\textgreater Q\textgreater Gm &
C\textgreater G\textgreater Q\textgreater Gm & yes \\
\end{tabular}
\end{table*}

Two observations upgrade the mechanism claim from ``consistent with'' to
``partially isolated'':

\begin{enumerate}
\def\labelenumi{\arabic{enumi}.}
\tightlist
\item
  A3 and A5 each individually break aggregator agreement. The
  \{A3\}-only and \{A5\}-only conditions both show disagreement among
  BT, Borda, and Ranked Pairs. Either identifiability-failing metric
  alone is sufficient to produce aggregator divergence on this dataset.
\item
  Removing either A3 or A5 is sufficient to restore agreement.
  Both LOMO-A3 and LOMO-A5 restore all-methods-agree, confirming that A3
  and A5 are the metrics on the full sweep that the aggregators are
  disagreeing about --- not A1 / A2 / A4.
\end{enumerate}

One counterintuitive finding is worth flagging:

\begin{enumerate}
\def\labelenumi{\arabic{enumi}.}
\setcounter{enumi}{2}
\tightlist
\item
  LOMO-A4 does NOT restore agreement --- removing the weighted-
  kappa metric (one of the three identifiability-passing metrics) while
  keeping A3 and A5 leaves the aggregators disagreeing. The
  interpretation is that A4 is load-bearing for consensus in the mixed
  sweep rather than a source of disagreement: its between-model signal
  stabilises BT against the noise injected by A3 and A5, and once A4 is
  removed the stabilisation is lost even though A1 and A2 are still
  present.
\end{enumerate}

These observations support the following claim, stronger than in earlier
drafts but still bounded:

\begin{quote}
On JF-ICR, removing either of the two identifiability-failing metrics
(A3 or A5) is \emph{individually} sufficient to restore aggregator
agreement; retaining either one individually is sufficient to break it.
Among the three passing metrics, A4 (weighted kappa) is load-bearing for
the consensus property and A1 / A2 are not. We claim this as a
decomposition of the aggregator-disagreement mechanism on this dataset,
not as a general property of identifiability failure.
\end{quote}

See §8.1 for the broader mechanism discussion.

\subsection{CI-significant inversions are
rare}\label{ci-significant-inversions-are-rare}

A naive count of ``policies on which model A ranks above model B''
inflates under per-policy noise. We formalise the test as follows.

Test definition. For each (policy k = (rubric, temperature),
ordered pair (a, b)) we compute an exact paired-randomisation p-value on
the per-item correctness difference indicator d\_i = 1{[}a correct on
item i{]} − 1{[}b correct on item i{]}: under the paired null, sign-flip
each d\_i independently, and the two-sided p-value is the
Phipson--Smyth-corrected Monte-Carlo estimate
\cite{ref-phipson2010} (B = 100 000 sign-flip draws, $\hat{p}$ = (r + 1)/(B +
1) where r is the count of \textbar Σ
sign⋅d\_i\textbar{} ≥ \textbar Σ d\_i\textbar; §7.4, completed). The
grid has 6 ordered pairs × 5 rubrics × 3 temperatures = 90
tests; there is no aggregator dimension because the per-item
correctness indicator is identical across A1, A2, and A4 (all three
reduce to the same 0 / 1 per-item statistic for this test). We apply
Holm--Bonferroni at α = 0.05 for the family-wise claim and
Benjamini--Hochberg at q = 0.05 for exploratory false-discovery control.
Aggregator-level significance (not per-item) is out of scope for this
test and is handled by the §7.2 paired-Δ CIs with Holm correction at m =
6.

\begin{table*}[!ht]
\centering
\small
\caption{Exact paired-randomisation test (B = 100,000, Phipson--Smyth)
on the 90-cell inversion grid. 41 of 90 cells reject at uncorrected
α = 0.05; the per-pair breakdowns below are for the family-wise and
FDR-controlled rejects, where per-pair stratification is
interpretable.}
\begin{tabular}{llll}
\toprule\noalign{}
Pair & n\_cells & Holm rejects (α=0.05) & BH-FDR rejects (q=0.05) \\
\midrule\noalign{}
claude vs gpt & 15 & 0 & 0 \\
claude vs gemini & 15 & 0 & 0 \\
claude vs qwen & 15 & 2 (R2 literal τ=0.3, 0.7) & 4 \\
gpt vs gemini & 15 & 0 & 1 \\
gpt vs qwen & 15 & 3 (R3 pragmatic τ=0.0, 0.3, 0.7) & 6 \\
gemini vs qwen & 15 & 0 & 6 \\
Total & 90 & 5 & 17 \\
\end{tabular}
\end{table*}

At uncorrected α = 0.05, 41 of the 90 cells (45.6\%) reject. Five
(5.6\%) survive Holm family-wise correction, and 17 (18.9\%) survive
BH-FDR. The Holm
rejects concentrate on pairs involving the weakest model (Qwen) ---
Claude-vs-Qwen (two rejects on R2 literal at nonzero temperature) and
GPT-vs-Qwen (three rejects on R3 pragmatic across all three
temperatures) --- consistent with §7.2's observation that Qwen sits
below the other three on aggregate accuracy at the primary cell. The B =
500 paired-bootstrap from earlier drafts resolved to minimum p = 4 ×
10⁻³, an order of magnitude above the Holm threshold α/m = 5.6 × 10⁻⁴
and could not produce any determinate rejections; the B = 100 000
Monte-Carlo permutation test has minimum resolvable p = 1/(B+1) ≈ 10⁻⁵,
well below the Holm threshold, and therefore gives determinate reject /
retain outcomes throughout the grid.

What the 5 Holm-surviving rejections say, and do not say. They
are per-policy rejections of the null that models a and b are equally
accurate on the exact-match indicator at that particular (rubric,
temperature). They do \emph{not} establish that models disagree across
policies, nor that any given ranking is stable under policy
perturbation. The aggregate-ranking conclusions in §7.2--§7.3 rest on
per-model CIs, pairwise Δ tests with Holm correction at m = 6, and the
aggregator-agreement observation on the identifiable-metric subset ---
not on this per-policy grid.

The grid has 90 cells, each computed with seed = 42 and B = 100 000
sign-flip draws.

\subsection{Dimension sensitivity
(descriptive)}\label{dimension-sensitivity-descriptive}

Definition. For each dimension d ∈ \{rubric, aggregation,
temperature\}, we compute the mean Kendall τ-distance \cite{ref-kendall1938}
between rankings produced by pairs of policies that differ only on d
(and agree on the other two dimensions), averaged over all such pairs in
the policy grid. τ-distance is normalised to {[}0, 1{]}, where 0 means
identical rankings and 1 means fully inverted.

\begin{table*}[!ht]
\centering
\small
\caption{Mean normalised Kendall τ-distance between per-policy rankings
across pairs that differ only along the named dimension. Larger = more
ranking movement attributable to that dimension.}
\begin{tabular}{lll}
\toprule\noalign{}
Dimension & Clean \{A1, A2, A4\} & Full \{A1..A5\} \\
\midrule\noalign{}
Rubric & 0.474 & 0.440 \\
Aggregation metric & 0.193 & 0.280 \\
Temperature & 0.119 & 0.120 \\
\end{tabular}
\end{table*}

These are point estimates reported descriptively. Uncertainty
quantification is not yet integrated: the clustered paired-bootstrap in
§7.2 covers per-policy scores and pairwise Δ, but re-scoring every
policy and recomputing the per-dimension mean τ-distance inside each
bootstrap resample remains outside the current uncertainty analysis
(§9.4). Because the
Kendall-τ decomposition does not yet carry an uncertainty band, we avoid
multiplicative-ratio language (``2.5 ×'', ``4 ×'') and treat this
subsection as descriptive rather than inferential.

A reader may note from the table that rubric appears to be the largest
single source of ranking movement in this grid, followed by aggregation,
followed by temperature. That point-estimate ordering is suggestive but
not tested; it enters the paper as a descriptive observation consistent
with the rubric-sensitivity finding in §7.1, not as an inferential
claim.

Moving from the full metric set to the clean subset reduces apparent
aggregation sensitivity from 0.280 to 0.193, a relative drop of ≈31\%.
The drop is a mechanical consequence of A3 and A5 failing the
identifiability rule in §5 --- a pre-specified failure, not a post-hoc
judgment about the metrics' usefulness.

\subsection{Condorcet analysis: a dominant winner, zero
cycles}\label{condorcet-analysis-a-dominant-winner-zero-cycles}

Applying Condorcet \cite{ref-condorcet1785} analysis over the 45-policy clean subset:

\begin{itemize}
\tightlist
\item
  Condorcet winner: claude-sonnet-4-6. Claude wins the pairwise
  majority against each of the other three models.
\item
  Condorcet cycles: zero. No three models form a non-transitive
  A \textgreater{} B \textgreater{} C \textgreater{} A preference
  pattern.
\end{itemize}

The data are consistent with a simple reading: one dominant top model
and a close lower cluster of three whose per-policy scores overlap in
confidence. This is \emph{evidence of capability tiering}, not evidence
of voting-theoretic pathology, and we take up its interpretation as a
boundary condition in §8.2.

Permutation test (formal specification). \emph{Null hypothesis
H₀}: per-policy model rankings are exchangeable across policies ---
equivalently, the assignment of pairwise wins to specific policies is a
random permutation of the observed wins. \emph{Test statistic} T: the
sum over all six ordered model pairs of the per-pair inversion rate
against the aggregate BT ranking (lower T = more policy-consistent
ranking). \emph{Null distribution}: obtained by permuting per-policy
pairwise win indicators independently within each pair and recomputing
T, over B\_perm = 10 000 permutations. \emph{Alternative}: two-sided,
reporting the tail in which the observed T lands.

\emph{Observed} T = 2.10; null mean T = 2.71; null 2.5th percentile =
2.31; null 97.5th percentile = 3.15; two-sided p \textless{} 0.01.

The observed T lies \emph{below} the null mean: rankings are
more consistent across policies than chance would produce.
Models leave a fingerprint through the policy grid. This is a test of
whether between-model capability differences exist at all --- they do
--- rather than a test of whether rankings are policy-sensitive in the
paper's sense. Both a policy-invariant signal (true model ordering) and
a policy-sensitive-but-correlated signal (varying per-policy scores but
consistently ordered) would reject this null. The permutation result is
therefore weaker evidence for policy sensitivity than its small-p-value
presentation alone suggests, and we interpret it accordingly.

\section{Discussion}\label{discussion}

\subsection{Aggregator disagreement and identifiability
failure}\label{aggregator-disagreement-and-identifiability-failure}

A common framing in the aggregation literature is that Bradley--Terry
\emph{hides} cyclic preferences by projecting them onto a single scalar
and that Condorcet-compliant methods reveal disagreement BT smooths
over. Our data sit within that framing as follows, with the causal
status of each observation made explicit.

Empirical pattern. Zero Condorcet cycles on our data --- a
clear Condorcet winner (Claude) and no non-transitive triples. On the
clean subset \{A1, A2, A4\} all three aggregators (BT, Borda, Ranked
Pairs) produce the same ranking. On the full five-metric sweep BT
disagrees with Borda and Ranked Pairs on positions 3 and 4, and the LOMO
decomposition (§7.3) shows that A3 alone is sufficient to break
consensus, A5 alone is sufficient to break consensus, and LOMO-A4 does
not restore consensus --- A4 is load-bearing rather than a source of
disagreement.

Two readings of the overall pattern. The data are consistent
with a measurement-failure reading: aggregator disagreement disappears
once identifiability-failing metrics are removed, so the disagreement
co-occurs with and is explained by identifiability failure. They are
also consistent with a boundary-condition reading: a capability-tiered
four-model set with one dominant winner and three close losers does not
exhibit cyclic preferences regardless of which metrics are in the sweep,
and the BT-vs-Borda tip on the Gemini/Qwen position is a near-tie
artefact. Our four-model, N = 253, single-dataset evidence does not
discriminate between these readings. We do not contradict published
intransitivity findings in broader LLM evaluation --- they are reported
under conditions (larger / more capability-heterogeneous model sets,
richer task spaces) that differ from ours in ways that make cycles
structurally more likely. We supply a case study at the opposite end of
the regime axis and flag confirmatory replication on a benchmark where
cycles have been reported as the discriminating experiment (§9.5).

Practitioner implication. On a benchmark where BT and
Condorcet-compliant methods disagree, run an identifiability diagnostic
over the metrics before concluding that the aggregators substantively
disagree on the underlying preference structure. Metric degeneracy under
class imbalance is at least a plausible alternative to social-choice
pathology, and on the task we audit it is empirically consistent with
the observed pattern.

\subsection{Implications for Financial NLP
Deployment}\label{implications-for-financial-nlp-deployment}

The pattern matters for applied financial-NLP governance. A ranking that
flips Gemini and Qwen on a downstream evaluation should not drive a
procurement or deployment decision without first checking whether the
flip survives a paired-bootstrap CI on the evaluation set and whether it
depends on a weakly identified metric. On a task like implicit
commitment recognition, where the production of a +1 vs 0 label can feed
an IR-monitoring dashboard, an analyst-sentiment index, or a
covenant-breach alerting pipeline, the cost of acting on a
measurement-fragile ranking can be real even if the benchmark has gold
labels.

Three deployment checkpoints follow from the audit:

\begin{enumerate}
\def\labelenumi{\arabic{enumi}.}
\tightlist
\item
  Before model procurement: assess whether vendor leaderboards
  are trustworthy by requiring rubric text, class distribution,
  per-metric diagnostics, and uncertainty intervals. A leaderboard that
  omits these details is not measurement-complete for a financial buyer.
\item
  Before model deployment: check whether the reported ranking
  is stable enough to support the deployment decision. Differences
  smaller than the per-policy CI width, or rankings that appear only
  under weakly identified metrics, should not be treated as operational
  evidence.
\item
  During model operation: keep audit records that explain why
  the evaluation evidence is, or is not, adequate for a commitment label
  to enter compliance, risk, or investor-relations workflows. The record
  should specify the rubric, metric, aggregation policy, benchmark
  artifact, and claim scope, not just the model name and headline score.
\end{enumerate}

These checkpoints are governance checks, not evidence that the evaluated
models improve investment, compliance, or disclosure outcomes. The paper
supports a prior question: whether the benchmark result is stable enough
to be used as decision evidence at all.

The case-study evidence transfers procedurally, not empirically. The
same audit discipline is most relevant for financial NLP benchmarks with
ordinal or asymmetric labels, class imbalance, expert-judgment rubrics,
and downstream use in procurement, monitoring, or compliance review.
Earnings-call stance labels, guidance-change labels,
disclosure-materiality tags, covenant-adjacent alerts, and ESG
commitment labels fit this risk profile. These examples are transfer
conditions, not additional validation; they identify where the audit
should be run before leaderboard claims are used.

More broadly, sensitivity to task formulation is not unique to financial
NLP: companion work on AI search also finds that observed system
behavior depends on how the task is operationalized \cite{ref-zhu2026}.
We use that wider metrology framing only as context. The evidence in this
paper is narrower: it localises measurement sensitivity to a supervised
Japanese financial NLP benchmark and to the specific pathway from rubric
wording, metric identifiability, and aggregation policy to ranking
claims.

\subsection{Claim scope}\label{claim-scope}

The table below separates the claims supported by this experiment from
claims that remain outside its evidentiary reach.

\begin{center}
\small
\refstepcounter{table}
\label{tab:claimscope}
Table \thetable: Claim scope for the JF-ICR measurement
audit.

\begin{tabular}{p{0.43\linewidth}p{0.43\linewidth}}
\toprule\noalign{}
Supported by this experiment & Not supported by this experiment \\
\midrule\noalign{}
Rubric wording materially changes model-assigned labels on JF-ICR. &
One rubric is definitively culturally correct. \\
Some metrics are weakly identified under this class distribution. &
The model ranking generalizes to all financial NLP tasks. \\
Ranking conclusions depend on evaluation policy. &
The method improves downstream investment or compliance decisions. \\
Supervised financial NLP benchmarks need rubric / metric audit. &
Japanese pragmatics are causally isolated from prompt verbosity and
examples. \\
\bottomrule
\end{tabular}
\end{center}

\section{Limitations and future work}\label{limitations-and-future-work}

\subsection{Linguistic validation limits}\label{linguistic-validation-limits}

The R2 and R3 rubrics differ in more than one way: semantics, examples,
and verbosity move together. R1-vs-R5 wording sensitivity is comparable
in magnitude to R2-vs-R3 framing sensitivity, so the experiment supports
rubric-sensitive model behaviour but does not causally isolate Japanese
pragmatics or validate one rubric as culturally correct. A factorial
framing × verbosity rubric design and bilingual financial-language
adjudication would be the appropriate future-work path for the stronger
linguistic claim.

\subsection{Dataset provenance limits}\label{dataset-provenance-limits}

The N = 94 versus N = 253 discrepancy is not corrected upstream as of
the 2026-04-28 access date. The live HF viewer and our local artifact
agree on 253 rows; the card / README text still states 94. All reported
results use the observed 253-row local artifact
\texttt{data/}\allowbreak\texttt{jf\_icr\_253.jsonl}, whose SHA-256 is
reported in §3.2 and whose parsing produced zero errors. If those 253
rows later prove to
include variants of a smaller independent-item set, the bootstrap unit
of analysis would need to be revised and confidence intervals recomputed.
This does not invalidate the reported 253-row audit; it bounds the
provenance claim attached to it and shows why benchmark artifacts need
checksums, access dates, row-level manifests, and explicit source-of-truth
rules when scores are used in financial AI governance.

\subsection{Generalization limits}\label{generalization-limits}

JF-ICR is one supervised, Japanese financial NLP benchmark with four
frontier classifiers and one imbalanced five-class distribution. The
audit does not establish a universal model ranking, a general theory of
LLM leaderboard instability, or downstream improvement in investment,
disclosure, or compliance decisions. It establishes that, on this
benchmark, rubric, metric, and aggregation policy are part of the
measurement instrument and should be reported as such. Its general claim
is procedural: supervised financial NLP benchmarks should expose enough
provenance, metric-identifiability, uncertainty, and aggregation detail
for reviewers and buyers to determine whether a ranking is
measurement-supported. What may generalize is the reporting discipline;
what does not generalize is the model ranking, the exact metric failures,
the JF-ICR class distribution, or the pragmatic-boundary interpretation.

\subsection{Uncertainty and repeatability
limits}\label{uncertainty-and-repeatability-limits}

N = 253 is modest for five-class ordinal classification. The
paired-bootstrap CI half-width on a per-policy accuracy at p ≈ 0.56 is
±6.1 pp, so small pairwise differences are not resolvable. Severe class
imbalance (2 strong-refusal, 11 weak-refusal items) limits metrics
anchored in minority classes. Same-policy repeatability across API seeds
is not yet measured, and the Kendall-τ dimension-sensitivity
decomposition remains descriptive because uncertainty is not integrated
inside that decomposition.\footnote{All four classifiers are run at
default effort with chain-of-thought / thinking features off. This keeps
the policy grid controlled, but means the per-model scores should not be
read as reasoning-enabled performance ceilings.}

\subsection{Future work}\label{future-work}

\begin{enumerate}
\def\labelenumi{\arabic{enumi}.}
\tightlist
\item
  Linguistic validation: run a 2 × 2 framing × verbosity rubric
  design and bilingual adjudication on a +1 / 0-enriched subset.
\item
  Repeatability: re-run the primary cell under 10 API seeds per
  model and report within-cell SD of A1/A2/A4, to quantify same-policy
  variance directly rather than infer it from the τ dimension.
\item
  Confirmatory replication: repeat the audit on an independent
  culturally
  grounded ordinal benchmark under a pre-registered protocol, and on a
  capability-heterogeneous model set where BT-vs-Condorcet disagreement
  has already been reported in the literature --- the natural
  discriminating test between the measurement-failure and
  boundary-condition readings of §8.1.
\end{enumerate}

\section{Conclusion}\label{conclusion}

The main result is not a new leaderboard. It is that a supervised
financial NLP benchmark with gold labels can still produce
policy-dependent evidence unless the evaluation instrument is audited.
The experiment is deliberately bounded: one task, one observed 253-row
test split, four frontier classifiers, five rubric variants, three
temperatures, and five ordinal metrics. Within those bounds, the central
lesson is straightforward: financial AI benchmark scores are measurement
outputs, not neutral facts. Before they are used for model selection or
deployment, the rubric, metric, aggregation rule, and benchmark artifact
need to be part of the reported evidence.

First, rubric wording materially changes model-assigned labels.
R2--R3 agreement ranges from 70.0\% to 83.4\%, and the dominant movement
is concentrated near the +1 / 0 implicit-commitment boundary. That
pattern is consistent with a pragmatic-boundary interpretation, but the
experiment does not causally isolate Japanese pragmatics from prompt
verbosity and examples.

Second, metric identifiability changes which ranking claims are
defensible. Within-one accuracy is weakly identified because near misses
receive credit and the majority class dominates. Worst-class accuracy is
weakly identified because the rarest class has only two examples. Exact
accuracy, macro-F1, and weighted κ are the primary ranking metrics under
the operational rule used here.

Third, aggregation conclusions depend on that metric audit.
Bradley--Terry, Borda, and Ranked Pairs agree on the identifiable metric
subset; the full five-metric sweep disagrees on the closest pair. The
leave-one-metric-out decomposition shows that A3 and A5 are sufficient
to break agreement on this dataset, while A4 helps stabilize it. This is
evidence for measurement risk in this supervised benchmark, not a
universal theory of LLM ranking behavior.

For financial NLP practice, the discipline is concrete: report the
rubric, pin the benchmark artifact, report metric-identifiability
diagnostics, report uncertainty, and state the claim scope. JF-ICR is
therefore the gold-labels-exist case for Financial AI Metrology Across
Ground-Truth Regimes. It audits the benchmark ruler when labels exist
and shows why measurement governance is necessary before the harder
pre-realization setting where gold labels are unavailable.

\appendix

\section{Metric degeneracy under label
imbalance}\label{metric-degeneracy-under-label-imbalance}

We report A3 (within-one accuracy) and A5 (worst-class accuracy) here
rather than in the main results because their behaviour on JF-ICR is
dominated by artefacts of the class distribution rather than by model
capability. They are not \emph{bad} metrics in general. They are
\emph{weakly identified} on this dataset's label counts, and the
distinction matters: on a balanced ordinal dataset, A3 and A5 would each
carry information that A1, A2, and A4 do not.

\subsection{A3 within-one accuracy
saturates}\label{a3-within-one-accuracy-saturates}

A3 awards credit when \textbar predicted − gold\textbar{} ≤ 1. On
JF-ICR, the predict-always-+1 baseline already achieves
\textasciitilde0.91 on A3, because +1 is the majority class and the
nearest-neighbour classes (0, +2) are within distance 1. The score range
across all 60 policies on A3 is 0.806--0.957, with per-model spreads of
0.040--0.111. The metric has essentially no discriminative headroom on
this label distribution. A ranking induced by A3 alone tracks variance
in ``approximate majority-class behaviour,'' not variance in full-scale
ordinal capability.

\subsection{A5 worst-class accuracy is dominated by two
samples}\label{a5-worst-class-accuracy-is-dominated-by-two-samples}

A5 takes the minimum per-class accuracy. On JF-ICR, class −2 contains 2
samples, so A5 on class −2 has three possible values: 0/2, 1/2, 2/2. A
single sample flip on the rarest class moves A5 by 0.5. Across the 15
policies on which A5 is the active metric, the score range is
0.000--0.364 with per-model spreads up to 0.36 --- nearly all of which
is a function of whether the model happens to get either of the two −2
samples right. A5 would be a legitimate headline metric on a balanced
ordinal dataset. At this imbalance, its per-policy score is
indistinguishable from sampling noise on the rarest class.

\subsection{Effect on dimension sensitivity and rank
aggregation}\label{effect-on-dimension-sensitivity-and-rank-aggregation}

Moving from the full metric set \{A1, A2, A3, A4, A5\} to the clean
subset \{A1, A2, A4\} reduces the estimated aggregation-metric
sensitivity from 0.280 to 0.193 Kendall τ-distance. Approximately
one-third of the apparent ``aggregation-metric sensitivity'' in the full
sweep is a consequence of A3 saturation and A5 sample-count dominance,
not genuine disagreement among well-identified metrics.

The effect extends to rank aggregation. On the 75-policy full sweep,
Bradley--Terry ranks the four models \{claude, gpt, qwen, gemini\} while
Borda and Ranked Pairs produce \{claude, gpt, gemini, qwen\} --- a swap
on the closest pair. On the 45-policy clean subset, all three methods
produce the same ranking. Rank-aggregation disagreement on the full
sweep is therefore attributable to A3 and A5 contributing
noise-dominated per-policy entries to the win matrix, which BT and the
Condorcet-compliant methods absorb differently. The clean-subset
analysis is the one we report as the primary finding. The full-sweep
analysis is retained here as a diagnostic illustration of how metric
choice can produce artefactual rank-aggregation disagreement.

\section{Rubric variants (full prompts)}\label{rubric-variants-full-prompts}

We reproduce the verbatim system prompts for all five rubric variants
below. Every variant shares two blocks: the 5-point ordinal scale and
the JSON output instruction. The variants differ only in the
\emph{role line} at the top and the \emph{classification guidance}
that follows the scale. We show the shared blocks once and then
enumerate each variant's unique text.

\subsection*{Shared ordinal scale (R1--R4)}

R1 through R4 all include the following five-label scale verbatim
under the heading \emph{Implicit Stance Scale}:

\begin{itemize}
\tightlist
\item +2 (Strong Commitment) --- Unambiguous, specific,
  time-bound commitment to a concrete action or outcome. The speaker
  pledges something definite.
\item +1 (Weak / Qualified Commitment) --- Intent expressed
  with conditions, caveats, timelines, or hedging. The speaker shows
  willingness but attaches uncertainty.
\item 0 (Neutral / Hedged) --- Neither commitment nor
  refusal. The speaker acknowledges the question without committing
  or declining. May restate facts or redirect.
\item -1 (Weak Refusal) --- Polite deflection without
  explicit denial. The speaker avoids direct commitment, often using
  formulaic expressions that pragmatically signal non-engagement.
\item -2 (Strong Refusal) --- Explicit refusal to answer,
  act, or commit. The speaker clearly declines using direct refusal
  language.
\end{itemize}

R5 replaces the scale with a one-line gloss per label (reproduced
under R5 below).

\subsection*{Shared output format (all rubrics)}

Every variant closes with an identical instruction to respond with
only a single JSON object containing three fields: \texttt{label}
(an integer in $\{-2,-1,0,+1,+2\}$), \texttt{confidence} (a float in
$[0.0, 1.0]$), and \texttt{reasoning} (a one- to two-sentence
explanation).

\subsection*{R1 balanced}

Role line. ``You are an expert annotator for the JF-ICR
(Japanese Financial Implicit Commitment Recognition) task. You
classify corporate responses to investor/analyst questions on an
ordinal commitment scale.''

Classification guidance. ``Consider BOTH the literal meaning
of the words AND the pragmatic intent in Japanese corporate
communication. Japanese business language frequently uses indirect
expressions; weigh the cultural context alongside what is explicitly
stated. When a response uses standard corporate hedging language
(e.g.\ \jpn{検討してまいります, 前向きに取り組む}), consider whether the
hedging is genuine qualification or a pragmatic deflection.''

\subsection*{R2 literal}

Role line. ``You are an annotator for the JF-ICR task.
Classify corporate responses based on their LITERAL, SURFACE-LEVEL
meaning.''

Classification guidance. ``Classify based strictly on what
the words explicitly say, NOT on cultural conventions or implied
meaning.
\begin{itemize}
\tightlist
\item If the response says \jpn{検討してまいります} (we will consider it),
  classify as +1 because the words express an intent to
  consider.
\item If the response says \jpn{前向きに} (positively/forward-looking),
  take it at face value as a positive signal.
\item Do NOT infer pragmatic refusal from polite language ---
  classify only what is stated.''
\end{itemize}

\subsection*{R3 pragmatic}

Role line. ``You are an expert in Japanese corporate
communication patterns, annotating the JF-ICR task. Classify
corporate responses based on their PRAGMATIC INTENT in Japanese
business culture.''

Classification guidance. ``In Japanese corporate
communication, many expressions carry conventional pragmatic meanings
that differ from their literal content:
\begin{itemize}
\tightlist
\item \jpn{検討させていただきます} (we will consider it) → typically a
  polite deflection (-1), NOT a genuine commitment to
  consider.
\item \jpn{前向きに検討} (will consider positively) → slightly more
  committed than plain \jpn{検討} but still often deflection
  (-1 or 0).
\item \jpn{善処します} (we will deal with it appropriately) → usually a
  non-answer that avoids commitment (0).
\item \jpn{お答えを控えさせていただきます} → explicit refusal using humble
  form (-2).
\item \jpn{～と認識しております} (we recognize that\ldots) → acknowledges
  without committing to action (0).
\end{itemize}
Prioritize pragmatic intent over literal meaning. A response that
sounds positive in translation may be a deflection in context.''

\subsection*{R4 strict}

Role line. ``You are a conservative annotator for the JF-ICR
task. Apply HIGH thresholds for Strong Commitment (+2) --- only
assign +2 when the commitment is specific, time-bound, and
unequivocal.''

Classification guidance.
\begin{itemize}
\tightlist
\item +2 requires: specific action + specific timeline + no
  qualifications. ``We plan to\ldots'' is NOT +2.
\item +1 is the default for anything that expresses intent
  but includes ANY caveat, condition, or uncertainty.
\item 0 for responses that acknowledge the question but
  redirect or defer without engaging the substance.
\item -1 for responses that use polite deflection language
  without explicit refusal.
\item -2 only for EXPLICIT refusal language (\jpn{控える,
  できない, 回答しかねる}).
\end{itemize}
``When in doubt between +2 and +1, choose +1. When in doubt between
+1 and 0, choose +1.''

\subsection*{R5 minimal}

R5 collapses the entire prompt: no role line, no cultural-context
guidance, and a one-line gloss in place of the scale. The full R5
system prompt is:

\begin{quote}
\small
``Classify the corporate response on this scale:
\begin{itemize}
\tightlist
\item +2: Clear, explicit commitment to specific action with
  timeline
\item +1: Qualified commitment with conditions or caveats
\item 0: Neutral --- neither commitment nor refusal
\item -1: Polite deflection, avoids direct commitment or
  refusal
\item -2: Explicit refusal to answer or act
\end{itemize}
[output-format block follows, identical across rubrics].''
\end{quote}

\section{Bootstrap protocol, assumptions, and
adjustments}\label{bootstrap-protocol-assumptions-and-adjustments}

\subsection{C.1 Resampling design}\label{c.1-resampling-design}

All confidence intervals reported in the main text use clustered
paired-bootstrap resampling over items. For each iteration, item
indices are sampled with replacement from the full 253-item dataset; the
\emph{same} sampled indices are applied across all models, all rubrics,
all temperatures, and all aggregation metrics, preserving both pairing
across models and the within-item dependency induced by shared prompts.

Iteration counts by statistic:

\begin{center}
\footnotesize
\setlength{\tabcolsep}{3pt}
\refstepcounter{table}
\label{tab:resampling-counts}
Table \thetable: Paired-bootstrap / permutation iteration counts by
statistic.

\begin{tabular}{@{}p{0.54\columnwidth}p{0.16\columnwidth}p{0.23\columnwidth}@{}}
\toprule\noalign{}
Statistic & B & Where reported \\
\midrule\noalign{}
Per-model A1/A2/A4 point estimates & 2 000 & §7.2 \\
Pairwise A1 Δ (6 pairs, primary cell) & 2 000 & §7.2 \\
Per-model rubric-pair agreement & 2 000 & §7.1 \\
Per-class swap-rate CIs & 2 000 & §7.1 \\
Pair-policy inversion test (90 cells) & 100 000 & §7.4 \\
Permutation null for T & 10 000 & §7.6 \\
\bottomrule
\end{tabular}
\end{center}

Per-policy score estimates, pairwise Δ estimates, rubric-swap
percentages, and inversion indicators are computed on the resampled
items. Two-sided 95\% percentile intervals are reported. Bootstrap
two-sided p-values in §7.2 and §7.4 are computed as 2 · min($\hat{F}$(0), 1 −
$\hat{F}$(0)), where $\hat{F}$ is the empirical CDF of the bootstrap Δ distribution.

\subsection{C.2 Assumptions and integration
scope}\label{c.2-assumptions-and-integration-scope}

The clustered paired-item bootstrap treats the 253 items as exchangeable
draws from a superpopulation of Japanese investor- relations Q\&A. The
resampling integrates over item-level sampling variance only. It does
not integrate over: (a) model stochasticity on a single policy
(§9.4 and §9.5), (b) rubric-authoring variance (one rubric author per
R1..R5), or (c) label noise in the gold annotation (no dual-annotator
record is available; §9.1). These are acknowledged sources of residual
uncertainty not folded into the reported CIs.

\subsection{C.3 Multiple-comparisons
procedure}\label{c.3-multiple-comparisons-procedure}

Two comparison families carry family-wise corrections in the main text:

\begin{itemize}
\tightlist
\item
  6 headline pairwise Δ on A1 (§7.2): Holm--Bonferroni at
  α=0.05 reported alongside raw. The smallest raw two-sided p (0.012,
  Claude vs Qwen) exceeds the Holm α\_(1) = 0.0083, so no pair survives.
\item
  90 pair-policy inversion tests (§7.4): exact
  paired-randomisation with B = 100 000 sign-flip draws and
  Phipson--Smyth correction. Five cells survive Holm--Bonferroni at
  α=0.05; 17 cells survive Benjamini--Hochberg FDR at q=0.05.
\end{itemize}

Benjamini--Hochberg FDR control \cite{ref-benjamini1995} at q=0.05
on the 90-test grid is reported as an exploratory complement to Holm
family-wise control. We regard Holm as the primary adjustment for
confirmatory reading and BH as a softer diagnostic for exploratory
reading; both are reported in full in §7.4.

\subsection{C.4 Reproducibility}\label{c.4-reproducibility}

All statistics in this paper are computed under a single random seed
(42) from a single analysis run; every numerical value cited in the
main text is derived from that run. Future-work items listed in §9
target uncertainty bounds on existing point estimates (BT θ CIs,
τ-distance CIs, same-policy repeatability); none affects the
paper's qualitative claims.

\bibliographystyle{IEEEtran}
\bibliography{references}

\end{document}